# Learning to Identify Regular Expressions that Describe Email Campaigns

Paul Prasse                                                          PRASSE@CS.UNI-POTSDAM.DE
Christoph Sawade                                                     SAWADE@CS.UNI-POTSDAM.DE
Niels Landwehr                                                       LANDWEHR@CS.UNI-POTSDAM.DE
Tobias Scheffer                                                      SCHEFFER@CS.UNI-POTSDAM.DE

University of Potsdam, Department of Computer Science, August-Bebel-Strasse 89, 14482 Potsdam, Germany

## Abstract

This paper addresses the problem of inferring a regular expression from a given set of strings that resembles, as closely as possible, the regular expression that a human expert would have written to identify the language. This is motivated by our goal of automating the task of postmasters of an email service who use regular expressions to describe and blacklist email spam campaigns. Training data contains batches of messages and corresponding regular expressions that an expert postmaster feels confident to blacklist. We model this task as a learning problem with structured output spaces and an appropriate loss function, derive a decoder and the resulting optimization problem, and a report on a case study conducted with an email service.

## 1. Introduction

Popular spam dissemination tools allow users to implement mailing campaigns by specifying simple grammars that serve as message templates. A grammar is disseminated to nodes of a bot net, the nodes create messages by instantiating the grammar at random. Email service providers can easily sample elements of new mailing campaigns by collecting messages in spam traps or by tapping into known bot nets. When messages from multiple campaigns are collected in a joint spam trap, clustering tools can separate the campaigns reliably (Haider & Scheffer, 2009). However, probabilistic cluster descriptions that use a bag-of-words representation incur the risk of false positives, and it

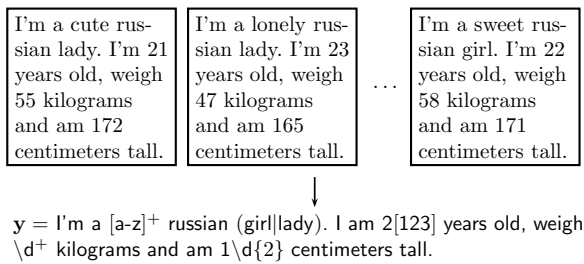

$\mathbf{y}$ = I'm a [a-z]$^+$ russian (girl|lady). I am 2[123] years old, weigh \d$^+$ kilograms and am 1\d{2} centimeters tall.

*Figure 1.* Elements of a message campaign and a regular expression created by a postmaster.

is difficult for a human to decide whether they in fact characterize the correct set of messages.

Regular expressions are a standard tool for specifying simple grammars. Widely available tools match strings against regular expressions efficiently and can be used conveniently from scripting languages. A regular expression can be translated into a finite state machine that accepts the language and has an execution time linear in the length of the input string. A specific, comprehensible regular expression which covers the observed instances and has been written by an expert postmaster can be used to blacklist the bulk of emails of that campaign at virtually no risk of covering any other messages.

Language identification has a rich history in the algorithmic learning theory community (see Section 6). Our problem setting differs from the problem of language identification in the learner's exact goal, and in the available training data. Batches of strings and corresponding regular expressions are observable in the training data. The learner's goal is to produce a predictive model that maps batches of strings to regular expressions that resemble as closely as possible the regular expressions which the postmaster would have written and feels confident to blacklist (see Figure 1).

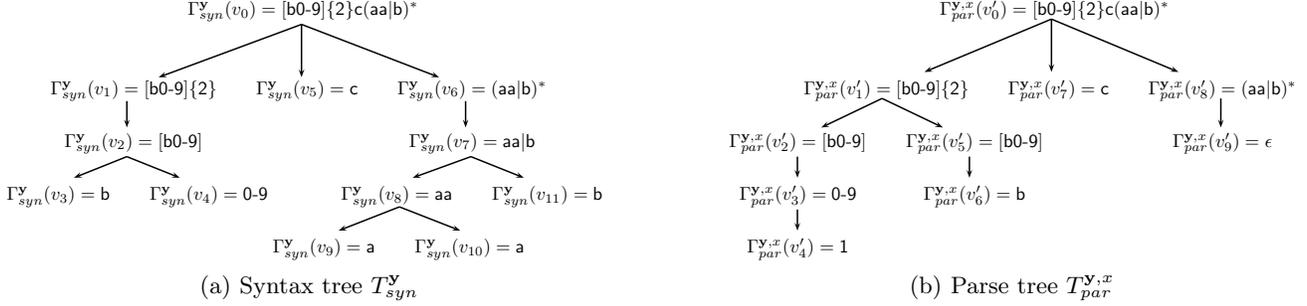

Figure 2. Syntax tree 2(a) and a parse tree 2(b) for the regular expression $\mathbf{y} = $ [b0-9]{2}c(aa|b)$^*$ and the string $x = $ 1bc.

The rest of this paper is structured as follows. Section 2 reviews regular expressions before Section 3 states the problem setting. Section 4 introduces the feature representation and derives the decoder and the optimization problem. In Section 5, we discuss our findings from a case study with an email service. Section 6 discusses related work; Section 7 concludes.

## 2. Regular Expressions

Syntactically, a regular expression $\mathbf{y} \in \mathcal{Y}_\Sigma$ is either a character from an alphabet $\Sigma$, or it is an expression in which an operator is applied to one or several argument expressions. Basic operators are the concatenation (*e.g.,* "abc"), disjunction (*e.g.,* "a|b"), and the *Kleene* star ("*"), written in postfix notation, that accepts any number of repetitions of its preceding argument expression. Parentheses define the syntactic structure of the expression. Several shorthands improve the readability of regular expressions and can be defined in terms of the basic operators. For instance, the *any character* symbol (".") abbreviates the disjunction of all characters in $\Sigma$, square brackets accept the disjunction of all characters (*e.g.,* "[abc]") or ranges (*e.g.,* "[a-z0-9]") that are included. The postfix operator "+" accepts an arbitrary, positive number of reiterations of the preceding expression, while "$\{l, u\}$" accepts between $l$ and $u$ reiterations, where $l \leq u$. We include a set of popular macros—for instance "\d" for *any digit*. A formal definition of the set of regular expressions can be found in the online appendix.

The syntactic structure of a regular expression is represented by its *syntax tree* $T_{syn}^{\mathbf{y}} = (V_{syn}^{\mathbf{y}}, E_{syn}^{\mathbf{y}}, \Gamma_{syn}^{\mathbf{y}}, \leq_{syn}^{\mathbf{y}})$. Definition 3 in the online appendix assigns one such tree to each regular expression. A node $v \in V_{syn}^{\mathbf{y}}$ of this tree is tagged by labeling function $\Gamma_{syn}^{\mathbf{y}} : V_{syn}^{\mathbf{y}} \to \mathcal{Y}_\Sigma$ with a subexpression $\Gamma_{syn}^{\mathbf{y}}(v) = \mathbf{y}_j$. Edges $(v, v') \in E_{syn}^{\mathbf{y}}$ indicate that node $v'$ represents an argument expression of $v$. Relation $\leq_{syn}^{\mathbf{y}} \subseteq V_{syn}^{\mathbf{y}} \times V_{syn}^{\mathbf{y}}$ defines an ordering on the nodes and identifies the root node.

A regular expression $\mathbf{y}$ defines a regular language $L(\mathbf{y})$. Given the regular expression, a deterministic finite state machine can decide whether a string $x$ is in $L(\mathbf{y})$ in time linear in $|x|$ (Dubé & Feeley, 2000). The trace of verification is typically represented as a *parse tree* $T_{par}^{\mathbf{y},x} = (V_{par}^{\mathbf{y},x}, E_{par}^{\mathbf{y},x}, \Gamma_{par}^{\mathbf{y},x}, \leq_{par}^{\mathbf{y},x})$, describing how the string $x$ can be derived from the regular expression $\mathbf{y}$. At least one parse tree exists if and only if the string is an element of the language $L(\mathbf{y})$; in this case, $\mathbf{y}$ is said to generate $x$. Nodes $v \in V_{syn}^{\mathbf{y}}$ of the syntax tree generate the nodes of the parse tree $v' \in V_{par}^{\mathbf{y},x}$; a node of the syntax tree may spawn none (alternatives which are not used to generate a string), one, or several ("loopy" syntactic elements such as "*" or "+") nodes in the parse tree. In analogy to the syntax trees, the labeling function $\Gamma_{par}^{\mathbf{y},x} : V_{par}^{\mathbf{y},x} \to \mathcal{Y}_\Sigma$ assigns a subexpression to each node, and the relation $\leq_{par}^{\mathbf{y},x} \subseteq V_{par}^{\mathbf{y},x} \times V_{par}^{\mathbf{y},x}$ defines the ordering of sibling nodes. The set of all parse trees for a regular expression $\mathbf{y}$ and a string $x$ is denoted by $\mathcal{T}_{par}^{\mathbf{y},x}$. A formal definition can be found in the online appendix.

Leaf nodes of a parse tree $T_{par}^{\mathbf{y},x}$ are labeled with elements of $\Sigma \cup \{\epsilon\}$, where $\epsilon$ denotes the empty symbol; reading them from left to right gives the generated string $x$. Non-terminal nodes correspond to subexpressions $\mathbf{y}_j$ of $\mathbf{y}$ which generate substrings of $x$. To compare different regular expressions with respect to a given string $x$, we define the set $T_{par}^{\mathbf{y},x}|_i$ of labels of nodes which are visited on the path from the root to the the $i$-th character of $x$ in the parse tree $T_{par}^{\mathbf{y},x}$.

Figure 2 shows an example of a syntax tree $T_{syn}^{\mathbf{y}}$ and a parse tree $T_{par}^{\mathbf{y},x}$ for the regular expression $\mathbf{y} = $ [b0-9]{2}c(aa|b)$^*$ and the string $x = $ 1bc.

Finally, we introduce the concept of a matching list. When a regular expression $\mathbf{y}$ generates a set $\mathbf{x}$ of strings, and $v \in V_{syn}^{\mathbf{y}}$ is an arbitrary node of the syn-tax tree of $\mathbf{y}$, then the matching list $M^{\mathbf{y},\mathbf{x}}(v)$ characterizes which substrings of the strings in $\mathbf{x}$ are generated by the node $v$ of the syntax tree. A node $v$ of the syntax tree generates a substring $x'$ of $x \in \mathbf{x}$, if $v$ generates a node $v'$ in the parse tree $T^{\mathbf{y},x}_{par}$ of $x$, and there is a path from $v'$ in that parse tree to every character in the substring $x'$. In the above example, for the set of strings $\mathbf{x} = \{\mathsf{12c}, \mathsf{b4ca}\}$, the matching list for node $v_1$ that represents subexpression [b0-9]{2} is $M^{\mathbf{y},\mathbf{x}}(v_2) = \{\mathsf{12}, \mathsf{b4}\}$. Definition 4 in the online appendix introduces matching lists formally.

## 3. Problem Setting

Having established the syntax and semantics of regular expressions, we now turn towards the problem setting. An unknown distribution $p(\mathbf{x}, \mathbf{y})$ generates regular expressions $\mathbf{y} \in \mathcal{Y}_\Sigma$ and batches $\mathbf{x}$ of strings $x \in \mathbf{x}$ that are elements of the language $L(\mathbf{y})$. In our motivating application, the strings $x$ are emails sampled from a bot net, and the $\mathbf{y}$ are regular expressions which an expert postmaster believes to identify the campaign template, and feels confident to blacklist.

A $\mathbf{w}$-parameterized predictive model $f_\mathbf{w} : \mathbf{x} \mapsto \hat{\mathbf{y}}$ accepts a batch of strings and conjectures a regular expression $\hat{\mathbf{y}}$. We now define the loss $\Delta(\mathbf{y}, \hat{\mathbf{y}}, \mathbf{x})$ that captures the deviation of the conjecture $\hat{\mathbf{y}}$ from $\mathbf{y}$ for batch $\mathbf{x}$. In our application, postmasters will not use an expression to blacklist the campaign unless they consider it to be comprehensibly and neatly written, and believe it to accurately identify the campaign.

Loss function $\Delta(\mathbf{y}, \hat{\mathbf{y}}, \mathbf{x})$ compares each of the accepting parse trees in $\mathcal{T}^{\mathbf{y},x}_{par}$, for each string $x \in \mathbf{x}$, with the most similar tree in $\mathcal{T}^{\hat{\mathbf{y}},x}_{par}$; if no such parse tree exists, the summand is defined as $\frac{1}{|\mathbf{x}|}$ (Equation 1). Similarly to a loss function for hierarchical classification (Cesa-Bianchi et al., 2006), the difference of two parse trees for string $x$ is quantified by a comparison of the paths that lead to the characters of the string; paths are compared by means of the intersection of their nodes (Equation 2). By its definition, this loss function is bounded between zero and one; it attains zero if and only if the expressions $\mathbf{y}$ and $\hat{\mathbf{y}}$ are equal.

$$\Delta(\mathbf{y}, \hat{\mathbf{y}}, \mathbf{x}) = \frac{1}{|\mathbf{x}|} \sum_{x \in \mathbf{x}} \begin{cases} \Delta_{\text{tree}}(\mathbf{y}, \hat{\mathbf{y}}, x) & \text{if } x \in L(\hat{\mathbf{y}}) \\ 1 & \text{otherwise} \end{cases} \quad (1)$$

with $\Delta_{\text{tree}}(\mathbf{y}, \hat{\mathbf{y}}, x)$ \hfill (2)

$$= 1 - \frac{1}{|\mathcal{T}^{\mathbf{y},x}_{par}|} \sum_{t \in \mathcal{T}^{\mathbf{y},x}_{par}} \max_{t' \in \mathcal{T}^{\hat{\mathbf{y}},x}_{par}} \frac{1}{|x|} \sum_{j=1}^{|x|} \frac{|t_{|j} \cap t'_{|j}|}{\max\{|t_{|j}|, |t'_{|j}|\}}$$

We will also explore the zero-one loss, $\Delta_{0/1}(\mathbf{y}, \hat{\mathbf{y}}, \mathbf{x}) = [\![\mathbf{y} \neq \hat{\mathbf{y}}]\!]$, where $[\![.]\!]$ is the indicator function of its boolean argument. The zero-one loss serves as an alternative, conceptually simpler reference model.

Our goal is to find the model $f_\mathbf{w}$ with minimal risk

$$R[f_\mathbf{w}] = \iint \Delta(\mathbf{y}, f_\mathbf{w}(\mathbf{x}), \mathbf{x}) p(\mathbf{x}, \mathbf{y}) \mathrm{d}\mathbf{x}\, \mathrm{d}\mathbf{y}. \quad (3)$$

Training data $D = \{(\mathbf{x}_i, \mathbf{y}_i)\}_{i=1}^m$ consists of pairs of batches $\mathbf{x}_i$ and generating regular expressions $\mathbf{y}_i$, drawn according to $p(\mathbf{x}, \mathbf{y})$.

Since the true distribution $p(\mathbf{x}, \mathbf{y})$ is unknown, the risk $R[f_\mathbf{w}]$ cannot be calculated. We state the learning problem as the problem of minimizing the regularized empirical counterpart of the risk over the parameters $\mathbf{w}$ and the regularizer $\Omega(\mathbf{w})$:

$$\hat{R}[f_\mathbf{w}] = \frac{1}{m} \sum_{(\mathbf{x},\mathbf{y}) \in D} \Delta(\mathbf{y}, f_\mathbf{w}(\mathbf{x}), \mathbf{x}) + \Omega(\mathbf{w}). \quad (4)$$

## 4. Identifying Regular Expressions

We model $f_\mathbf{w}$ as a linear discriminant function $\mathbf{w}^\mathsf{T} \Psi(\mathbf{x}, \mathbf{y})$ for a joint feature representation of the input $\mathbf{x}$ and output $\mathbf{y}$ (Tsochantaridis et al., 2005):

$$f_\mathbf{w}(\mathbf{x}) = \arg\max_{\mathbf{y} \in \mathcal{Y}_\Sigma} \mathbf{w}^\mathsf{T} \Psi(\mathbf{x}, \mathbf{y}). \quad (5)$$

### 4.1. Joint Feature Representation

The joint feature representation $\Psi(\mathbf{y}, \mathbf{x})$ captures structural properties of an expression $\mathbf{y}$ and joint properties of input batch $\mathbf{x}$ and regular expression $\mathbf{y}$.

Structural properties of a regular expression $\mathbf{y}$ are captured by features that indicate a specific nesting of regular expression operators—for instance, whether a concatenation occurs within a disjunction. More formally, we first define a binary vector

$$\Lambda(\mathbf{y}) = \begin{pmatrix} [\![\mathbf{y} = \mathbf{y}_1 \ldots \mathbf{y}_k]\!] \\ [\![\mathbf{y} = \mathbf{y}_1 | \ldots | \mathbf{y}_k]\!] \\ [\![\mathbf{y} = [\mathbf{y}_1 \ldots \mathbf{y}_k]]\!] \\ [\![\mathbf{y} = \mathbf{y}_1^*]\!] \\ [\![\mathbf{y} = \mathbf{y}_1?]\!] \\ [\![\mathbf{y} = \mathbf{y}_1^+]\!] \\ [\![\mathbf{y} = \mathbf{y}_1\{l\}]\!] \\ [\![\mathbf{y} = \mathbf{y}_1\{l,u\}]\!] \\ [\![\mathbf{y} = r_1]\!] \\ \vdots \\ [\![\mathbf{y} = r_l]\!] \\ [\![\mathbf{y} \in \Sigma]\!] \\ [\![\mathbf{y} = \epsilon]\!] \end{pmatrix} \quad (6)$$

encoding the top-level operator used in the regular expression $\mathbf{y}$. In Equation 6, $\mathbf{y}_1, \ldots, \mathbf{y}_k \in \mathcal{Y}_\Sigma$ are regular expressions, $l, u \in \mathbb{N}$, and $\{r_1, \ldots, r_l\}$ is a set of ranges and popular macros; for our application, we use the set $\{$0-9, a-f, a-z, A-F, A-Z, \S, \e, \w, \d, "."$\}$. For any two nodes $v'$ and $v''$ in the syntax tree of $\mathbf{y}$ that are connected by an edge—indicating that $\mathbf{y}'' = \Gamma_{syn}^{\mathbf{y}}(v'')$ is an argument subexpression of $\mathbf{y}' = \Gamma_{syn}^{\mathbf{y}}(v')$—the tensor product $\Lambda(\mathbf{y}') \otimes \Lambda(\mathbf{y}'')$ defines a binary vector that encodes the specific nesting of operators at node $v'$. Feature vector $\Psi(\mathbf{x}, \mathbf{y})$ will aggregate these vectors over all pairs of adjacent nodes in the syntax tree of $\mathbf{y}$.

Joint properties of an input batch $\mathbf{x}$ and a regular expression $\mathbf{y}$ are encoded as follows. Recall that for any node $v'$ in the syntax tree, $M^{\mathbf{y},\mathbf{x}}(v')$ denotes the set of substrings in $\mathbf{x}$ that are generated by the subexpression $\mathbf{y}' = \Gamma_{syn}^{\mathbf{y}}(v')$ that $v'$ is labeled with. We define a vector $\Phi(M^{\mathbf{y},\mathbf{x}}(v'))$ of attributes of this set. Any property may be accounted for; for our application, we include the average string length, the inclusion of the empty string, the proportion of capital letters, and many other attributes. The full list of attributes used in our experiments is included in the online appendix. A joint encoding of properties of the subexpression $\mathbf{y}'$ and the set of substrings generated by $\mathbf{y}'$ is given by the tensor product $\Phi(M^{\mathbf{y},\mathbf{x}}(v')) \otimes \Lambda(\mathbf{y}')$.

The joint feature vector $\Psi(\mathbf{x}, \mathbf{y})$ is obtained by aggregating operator-nesting information over all edges in the syntax tree, and joint properties of subexpressions $\mathbf{y}'$ and the set of substrings they generate over all nodes in the syntax tree:

$$\Psi(\mathbf{x}, \mathbf{y}) \quad (7)$$
$$= \begin{pmatrix} \sum_{(v',v'') \in E_{syn}^{\mathbf{y}}} \Lambda(\Gamma_{syn}^{\mathbf{y}}(v')) \otimes \Lambda(\Gamma_{syn}^{\mathbf{y}}(v'')) \\ \sum_{v' \in V_{syn}^{\mathbf{y}}} \Phi(M^{\mathbf{y},\mathbf{x}}(v')) \otimes \Lambda(\Gamma_{syn}^{\mathbf{y}}(v')) \end{pmatrix}.$$

### 4.2. Decoding

At application time, the highest-scoring regular expression $f_\mathbf{w}(\mathbf{x}) = \arg\max_{\mathbf{y} \in \mathcal{Y}_\Sigma} \mathbf{w}^\mathsf{T} \Psi(\mathbf{x}, \mathbf{y})$ has to be identified. This maximization is over the infinite space of all regular expressions $\mathcal{Y}_\Sigma$. To alleviate the intractability of this problem, we approximate this maximum by the maximum over a constrained, finite search space which can be found efficiently.

The constrained search space initially contains an alignment of all strings in $\mathbf{x}$. An alignment is a regular expression that contains only constants—which have to occur in all strings of the batch—and the wildcard symbol "(.*)". The initial alignment $\mathbf{a}_\mathbf{x}$ of $\mathbf{x}$ can be thought of as the most-general bound of this space.

**Definition 1** (Alignment). *The set of alignments $A_\mathbf{x}$ of a batch of strings $\mathbf{x}$ contains all concatenations in which strings from $\Sigma^+$ and the wildcard symbol "(.*)" alternate, and that generate all elements of $\mathbf{x}$.*

An alignment is maximal if no other alignment in $A_\mathbf{x}$ contains more constant symbols. A maximal alignment of two strings can be determined efficiently using Hirschberg's algorithm (Hirschberg, 1975) which is an instance of dynamic programming. By contrast, finding the maximal alignment of a *set of strings* is NP-hard (Wang & Jiang, 1994); known algorithms are exponential in the number $|\mathbf{x}|$ of strings in $\mathbf{x}$. *Progressive alignment* heuristics find an alignment of a set of strings by incrementally aligning pairs of strings.

Given an alignment $\mathbf{a}_\mathbf{x} = a_0(.^*)a_1 \ldots (.^*)a_n$ of all strings in $\mathbf{x}$, the constrained search space

$$\hat{\mathcal{Y}}_{\mathbf{x},D} = \{a_0 \mathbf{y}_1 a_1 \ldots \mathbf{y}_n a_n | \mathbf{y}_j \in \hat{\mathcal{Y}}_D^{M_j}\} \quad (8)$$

contains all specializations of $\mathbf{a}_\mathbf{x}$ in which the $j$-th wildcard symbol is replaced by any element of a set $\hat{\mathcal{Y}}_D^{M_j}$. The sets $\hat{\mathcal{Y}}_D^{M_j}$ are constructed by Algorithm 1. The algorithm starts with $\mathcal{Y}_D$ which we define to be the set of all subexpressions that occur anywhere in the training data $D$. From this set, it takes a subset such that each regular expression in $\hat{\mathcal{Y}}_{\mathbf{x},D}$ generates all strings in $\mathbf{x}$, and adds a number of syntactic variants and subexpressions in which constants have been replaced to match the elements of $M_j$, where $M_j$ is the matching list of the node which belongs to the $j$-th wildcard symbol. Each of the lines 7, 9, 10, 11, and 12 of Algorithm 1 adds at most one element to $\hat{\mathcal{Y}}_D^{M_j}$— hence, the search space of possible substitutions for each of the $n$ wildcard symbols is linear in the number of subexpressions that occur in the training sample.

We now turn towards the problem of determining the highest-scoring regular expression $f_\mathbf{w}(\mathbf{x})$. Maximization over all regular expressions is approximated by maximization over the space defined by Equation 8:

$$\arg\max_{\mathbf{y} \in \mathcal{Y}_\Sigma} \mathbf{w}^\mathsf{T} \Psi(\mathbf{x}, \mathbf{y}) \approx \arg\max_{\mathbf{y} \in \hat{\mathcal{Y}}_{\mathbf{x},D}} \mathbf{w}^\mathsf{T} \Psi(\mathbf{x}, \mathbf{y}). \quad (9)$$

We will now argue that this maximization problem can be decomposed into independent maximization problems for each of the $\mathbf{y}_j$ that replaces the $j$-th wildcard in the alignment $\mathbf{a}_\mathbf{x}$ due to the simple syntactic structure of the alignment and the definition of $\Psi$.

Feature vector $\Psi(\mathbf{x}, \mathbf{y})$ decomposes linearly into a sum over the nodes and a sum over pairs of adjacent nodes (see Equation 7). The syntax tree of an instantiation $\mathbf{y} = a_0 \mathbf{y}_1 a_1 \ldots \mathbf{y}_n a_n$ of the alignment $\mathbf{a}_\mathbf{x}$ consists of a root node labeled as an alternating concatenation of constant strings $a_j$ and subexpressions $\mathbf{y}_j$ (see Figure 3). This root node is connected to a layer on which constant strings $a_j = a_{j,1} \ldots a_{j,|a_j|}$ and subtrees $T_{syn}^{\mathbf{y}_j}$ alternate (blue area in Figure 3). However, the terms in Equation 10 that correspond to the root node $\mathbf{y}$ and

**Algorithm 1** Constructing the decoding space

**Input:** Subexpressions $\mathcal{Y}_D$ and alignment $\mathbf{a_x} = a_0(.^*)a_1\ldots(.^*)a_n$ of the strings in $\mathbf{x}$.
1: **let** $T^{\mathbf{a_x}}_{syn}$ be the syntax tree of the alignment and $v_1,\ldots,v_n$ be the nodes labeled $\Gamma^{\mathbf{a_x}}_{syn}(v_j) = $ "$(.^*)$".
2: **for** $j = 1\ldots n$ **do**
3:    **let** $M_j = M^{\mathbf{a_x},\mathbf{x}}(v_j)$.
4:    Initialize $\hat{\mathcal{Y}}_D^{M_j}$ to $\{\mathbf{y} \in \mathcal{Y}_D | M_j \subseteq L(\mathbf{y})\}$
5:    **let** $x_1,\ldots,x_m$ be the elements of $M_j$; add $(x_1|\ldots|x_m)$ to $\hat{\mathcal{Y}}_D^{M_j}$.
6:    **let** $u$ be the length of the longest string and $l$ be the length of the shortest string in $M_j$.
7:    **if** $[\beta\mathbf{y}_1\ldots\mathbf{y}_k] \in \hat{\mathcal{Y}}_D^{M_j}$, where $\beta \in \Sigma^*$ and $\mathbf{y}_1\ldots\mathbf{y}_k$ are ranges or special macros (e.g., a-z, \e), then add $[\alpha\mathbf{y}_1\ldots\mathbf{y}_k]$ to $\hat{\mathcal{Y}}_D^{M_j}$, where $\alpha \in \Sigma^*$ is the longest string that satisfies $M_j \subseteq L([\alpha\mathbf{y}_1\ldots\mathbf{y}_k])$, if such an $\alpha$ exists.
8:    **for all** $[\mathbf{y}] \in \hat{\mathcal{Y}}_D^{M_j}$ **do**
9:      add $[\mathbf{y}]^*$ and $[\mathbf{y}]\{l,u\}$ to $\hat{\mathcal{Y}}_D^{M_j}$.
10:      **if** $l = u$, then add $[\mathbf{y}]\{l\}$ to $\hat{\mathcal{Y}}_D^{M_j}$.
11:      **if** $u \leq 1$, then add $[\mathbf{y}]?$ to $\hat{\mathcal{Y}}_D^{M_j}$.
12:      **if** $l > 0$, then add $[\mathbf{y}]^+$ to $\hat{\mathcal{Y}}_D^{M_j}$.
13:    **end for**
14: **end for**
**Return:** $\hat{\mathcal{Y}}_D^{M_1},\ldots,\hat{\mathcal{Y}}_D^{M_n}$.

the $a_j$ are constant for all values of the $\mathbf{y}_j$ (red area in Figure 3). Since no edges connect multiple wildcards, the feature representation of these subtrees can be decomposed into $n$ independent summands as in Equation 11.

$$\Psi(\mathbf{x}, a_0\mathbf{y}_1 a_1\ldots\mathbf{y}_n a_n) \quad (10)$$

$$= \begin{pmatrix} \sum_{j=1}^n \Lambda(\mathbf{y}) \otimes \Lambda(\mathbf{y}_j) + \sum_{j=0}^n \sum_{q=1}^{|a_j|} \Lambda(\mathbf{y}) \otimes \Lambda(a_{j,q}) \\ \Phi(\{\mathbf{x}\}) \otimes \Lambda(\mathbf{y}) + \sum_{j=0}^n \sum_{q=1}^{|a_j|} \Phi(\{a_{j,q}\}) \otimes \Lambda(a_{j,q}) \end{pmatrix}$$

$$+ \begin{pmatrix} \sum_{j=1}^n \sum_{(v',v'')\in E^{\mathbf{y}_j}_{syn}} \Lambda(\Gamma^{\mathbf{y}_j}_{syn}(v')) \otimes \Lambda(\Gamma^{\mathbf{y}_j}_{syn}(v'')) \\ \sum_{j=1}^n \sum_{v' \in V^{\mathbf{y}_j}_{syn}} \Phi(M^{\mathbf{y}_j, M_j}(v')) \otimes \Lambda(\Gamma^{\mathbf{y}_j}_{syn}(v')) \end{pmatrix}$$

$$= \begin{pmatrix} \mathbf{0} \\ \Phi(\{\mathbf{x}\}) \otimes \Lambda(\mathbf{y}) \end{pmatrix} + \sum_{j=0}^n \sum_{q=1}^{|a_j|} \begin{pmatrix} \Lambda(\mathbf{y}) \otimes \Lambda(a_{j,q}) \\ \Phi(\{a_{j,q}\}) \otimes \Lambda(a_{j,q}) \end{pmatrix}$$

$$+ \sum_{j=1}^n \left( \Psi(\mathbf{y}_j, M_j) + \begin{pmatrix} \Lambda(\mathbf{y}) \otimes \Lambda(\mathbf{y}_j) \\ \mathbf{0} \end{pmatrix} \right) \quad (11)$$

Since the top-level operator of an alignment is a concatenation for any $\mathbf{y} \in \hat{\mathcal{Y}}_{\mathbf{x},D}$, we can write $\Lambda(\mathbf{y})$ as a constant $\Lambda_\bullet$, defined as the output feature vector (Equation 6) of a concatenation.

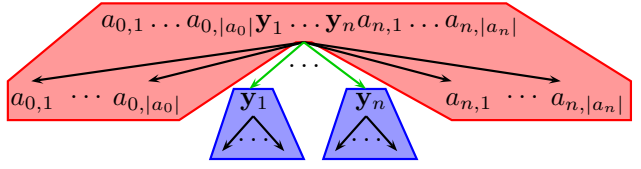

Figure 3. Structure of a syntax tree for an element of $\hat{\mathcal{Y}}_{\mathbf{x},D}$.

Thus, the maximization over all $\mathbf{y} = a_0\mathbf{y}_1 a_1 \ldots \mathbf{y}_n a_n$ can be decomposed into $n$ maximization problems over

$$\mathbf{y}_j^* = \arg\max_{\mathbf{y}_j \in \hat{\mathcal{Y}}_D^{M_j}} \mathbf{w}^\top \left( \Psi(\mathbf{y}_j, M_j) + \begin{pmatrix} \Lambda_\bullet \otimes \Lambda(\mathbf{y}_j) \\ \mathbf{0} \end{pmatrix} \right)$$

which can be solved in $\mathcal{O}(n \times |\mathcal{Y}_D|)$.

### 4.3. Optimization Problem

We will now address the process of minimizing the regularized empirical risk $\hat{R}$, defined in Equation 4, for the $\ell_2$ regularizer $\Omega(\mathbf{w}) = \frac{1}{2C}||\mathbf{w}||^2$. Loss function $\Delta$, defined in Equation 1, is not convex. To obtain a convex optimization problem, we upper-bound the loss by its hinged version, following the margin-rescaling approach (Tsochantaridis et al., 2005):

$$\xi_i = \max_{\mathbf{y} \neq \mathbf{y}_i}\{\mathbf{w}^\top(\Psi(\mathbf{x}_i, \mathbf{y}_i) - \Psi(\mathbf{x}_i, \mathbf{y})) + \Delta(\mathbf{y}_i, \mathbf{y}, \mathbf{x})\}. \quad (12)$$

The maximum in Equation 12 is over all $\mathbf{y} \in \mathcal{Y}_\Sigma \setminus \{\mathbf{y}_i\}$. When the risk is rephrased as a constrained optimization problem, the maximum produces one constraint per element of $\mathbf{y} \in \mathcal{Y}_\Sigma \setminus \{\mathbf{y}_i\}$. However, since the decoder searches only the set $\hat{\mathcal{Y}}_{\mathbf{x}_i, D}$, it is sufficient to enforce the constraints on this subset.

When the loss is replaced by its upper bound—the slack variable $\xi$—and for $\Omega(\mathbf{w}) = \frac{1}{2C}||\mathbf{w}||^2$, the minimization of the regularized empirical risk (Equation 4) is reduced to Optimization Problem 1.

**Optimization Problem 1.** *Over parameters $\mathbf{w}$, find*

$$\mathbf{w}^* = \arg\min_{\mathbf{w},\xi} \frac{1}{2}||\mathbf{w}||^2 + \frac{C}{m}\sum_{i=1}^m \xi_i, \text{ such that} \quad (13)$$

$$\forall i, \forall \bar{\mathbf{y}} \in \hat{\mathcal{Y}}_{\mathbf{x}_i,D}\setminus\{\mathbf{y}_i\}: \mathbf{w}^\top(\Psi(\mathbf{x}_i,\mathbf{y}_i) - \Psi(\mathbf{x}_i,\bar{\mathbf{y}})) \quad (14)$$
$$\geq \Delta(\mathbf{y}_i, \bar{\mathbf{y}}, \mathbf{x}) - \xi_i,$$

$$\forall i: \xi_i \geq 0. \quad (15)$$

This optimization problem is convex, since the objective (Equation 13) is convex and the constraints

(Equation 14 and 15) are affine in $\mathbf{w}$. Hence, the solution is unique and can be found efficiently by cutting plane methods as Pegasos (Shalev-Shwartz et al., 2011) or SVM$^{struct}$ (Tsochantaridis et al., 2005).

---

**Algorithm 2** Most strongly violated constraint

**Input:** batch $\mathbf{x}$, model $f_\mathbf{w}$, correct output $\mathbf{y}$.
1: Infer alignment $\mathbf{a_x} = a_0(.^*)a_1\ldots(.^*)a_n$ for $\mathbf{x}$.
2: Let $T^{\mathbf{a_x}}_{syn}$ be the syntax tree of $\mathbf{a_x}$ and let $v_1,\ldots,v_n$ be the nodes labeled $\Gamma^{\mathbf{a_x}}_{syn}(v_j) = \text{``}(.^*)\text{''}$.
3: **for all** $j = 1\ldots n$ **do**
4:    Let $M_j = M^{\mathbf{a_x},\mathbf{x}}(v_j)$ and calculate the $\hat{\mathcal{Y}}^{M_j}_D$ using Algorithm 1.
5:    $\bar{\mathbf{y}}_j = \displaystyle\arg\max_{\mathbf{y}'_j \in \hat{\mathcal{Y}}^{M_j}_D} \mathbf{w}^\mathsf{T}\left(\Psi(\mathbf{y}'_j, M_j) + \begin{pmatrix}\Lambda_\bullet \otimes \Lambda(\mathbf{y}'_j)\\ \mathbf{0}\end{pmatrix}\right) +$
    $\Delta(\mathbf{y},a_0(.^*)a_1\ldots(.^*)a_{j-1}\mathbf{y}'_j a_j(.^*)a_{j+1}\ldots(.^*)a_n, \mathbf{x})$
6: **end for**
7: Let $\bar{\mathbf{y}}$ abbreviate $a_0\bar{\mathbf{y}}_1 a_1 \ldots \bar{\mathbf{y}}_n a_n$
8: **if** $\bar{\mathbf{y}} = \mathbf{y}$ **then**
9:    Assign a value of $\bar{\mathbf{y}}'_j \in \hat{\mathcal{Y}}^{M_j}_D$ to one of the variables $\bar{\mathbf{y}}_j$ such that the smallest decrease of $f_\mathbf{w}(\mathbf{x}, \bar{\mathbf{y}}) + \Delta_\text{tree}(\mathbf{y}, \bar{\mathbf{y}})$ is obtained but the constraint $\bar{\mathbf{y}} \neq \mathbf{y}$ is enforced.
10: **end if**
**Return:** $\bar{\mathbf{y}}$

---

During the optimization procedure, the regular expression that incurs the highest slack $\xi_i$ for a given $\mathbf{x}_i$,

$$\bar{\mathbf{y}} = \arg\max_{\mathbf{y} \in \hat{\mathcal{Y}}_{\mathbf{x}_i,D} \setminus \{\mathbf{y}_i\}} \mathbf{w}^\mathsf{T}\Psi(\mathbf{x}_i, \mathbf{y}) + \Delta(\mathbf{y}_i, \mathbf{y}, \mathbf{x}),$$

has to be identified repeatedly. Algorithm 1 constructs the constrained search space $\hat{\mathcal{Y}}_{\mathbf{x}_i,D}$ such that $x \in L(\mathbf{y})$ for each $x \in \mathbf{x}_i$ and $\mathbf{y} \in \hat{\mathcal{Y}}_{\mathbf{x}_i,D}$. Hence, the "otherwise"-case in Equation 1 never applies within our search space. Without this case, Equations 1 and 2 decompose linearly over the nodes of the parse tree, and therefore the wildcards. Hence, $\bar{\mathbf{y}}$ can be identified by maximizing over the variables $\bar{\mathbf{y}}_j$ independently in Step 5 of Algorithm 2. Algorithm 2 finds the constraint that is violated most strongly within the constrained search space in $\mathcal{O}(n \times |\mathcal{Y}_D|)$. This ensures a polynomial execution time of the optimization algorithm. We refer to this learning procedure as *REx-SVM*.

## 5. Case Study

We investigate whether postmasters accept the output of *REx-SVM* to blacklist mailing campaigns during regular operations of a commercial email service. We also evaluate how accurately *REx-SVM* and reference methods identify the extensions of mailing campaigns.

| | Campaign 1 | Campaign 2 | Campaign 3 |
|---|---|---|---|
| Postmaster | First name: [ \S]$^+$<br>Surname: \S$^+$<br>Height: 1\d$^+$ cm.<br>Weights: \d{2} kg. | The trans(fer\|action)<br>ID: \d$^+\ldots$<br>ID:( )$^*$\d$^+$( )$^*\ldots$<br>report_\d$^+$.doc | http://(LOVEGAME<br>[S0-9]$^*$\|lovegame<br>[s0-9]$^*$).(com\|net) |
| REx-SVM | First name: [ \S]$^+$<br>Surname: \S$^+$<br>Height: 1\d$^+$ cm.<br>Weights: \d{2} kg. | The trans(fer\|action)<br>ID: \d$^+\ldots$<br>ID:[ 0-9]$^+\ldots$<br>report_\d$^+$.doc | http://\e$^+$.(com\|net) |
| REx$_{0/1}$-SVM | First name: [ \S]$^+$<br>Surname: [a-zA-Z]$^+$<br>Height: 1\d$^+$ cm.<br>Weights: [1467]$^+$ kg. | The trans[a-z]$^+$<br>ID: \d$^+\ldots$<br>ID:[ a-z0-9]{2,6}$\ldots$<br>report_(2\|\ldots\|73).doc | http://\e$^+$.[a-z]$^+$ |

*Figure 4.* Regular expressions created by a postmaster and corresponding output of *REx-SVM* and *REx$_{0/1}$-SVM*.

### 5.1. Evaluation by Postmasters

*REx-SVM* is trained on the *ESP* data set that contains 158 batches with a total of 12,763 emails and corresponding regular expressions, collected from the email service provider. The model is deployed; the user interface presents newly detected batches of spam emails together with the regular expression conjectured by *REx-SVM* to a postmaster during regular operations of the service. The postmaster is charged with blacklisting the campaigns by suitable regular expressions. Over the study, the postmasters created 188 regular expressions. Of these, they created 169 expressions (89%) by copying a substring of the automatically generated expression. We observe that postmasters prefer to describe only a part of the message which they feel is characteristic for the campaign whereas *REx-SVM* describes the entirety of the messages. In 12 cases, the postmasters edited the string, and in 7 cases they wrote an expression from scratch.

To illustrate different cases, Figure 4 compares excerpts of expressions created by *REx-* and *REx$_{0/1}$-SVM* (a variant of *REx-SVM* that uses the zero-one loss instead of $\Delta$ defined in Equation 1) to expressions of a postmaster. The first example shows a perfect agreement between *REx-SVM* and postmaster. In the second example, the expressions are close but distinct. In the third example, the SVMs produce expressions that generate an overly general set of URLs and lead to false positives ("\e" stands for characters that can occur in a URL). In all three cases, *REx-SVM* is more similar to the postmaster than *REx$_{0/1}$*.

The top right diagram of Figure 5 shows the average loss $\Delta$ of *REx-* and *REx$_{0/1}$-SVM*, measured by cross validation with one batch held out. While postmasters show the tendency to write expressions that only characterize about 10% of the message, the *REx-SVM* variants describe the entirety of the message. This leads to relatively high values of the loss function.

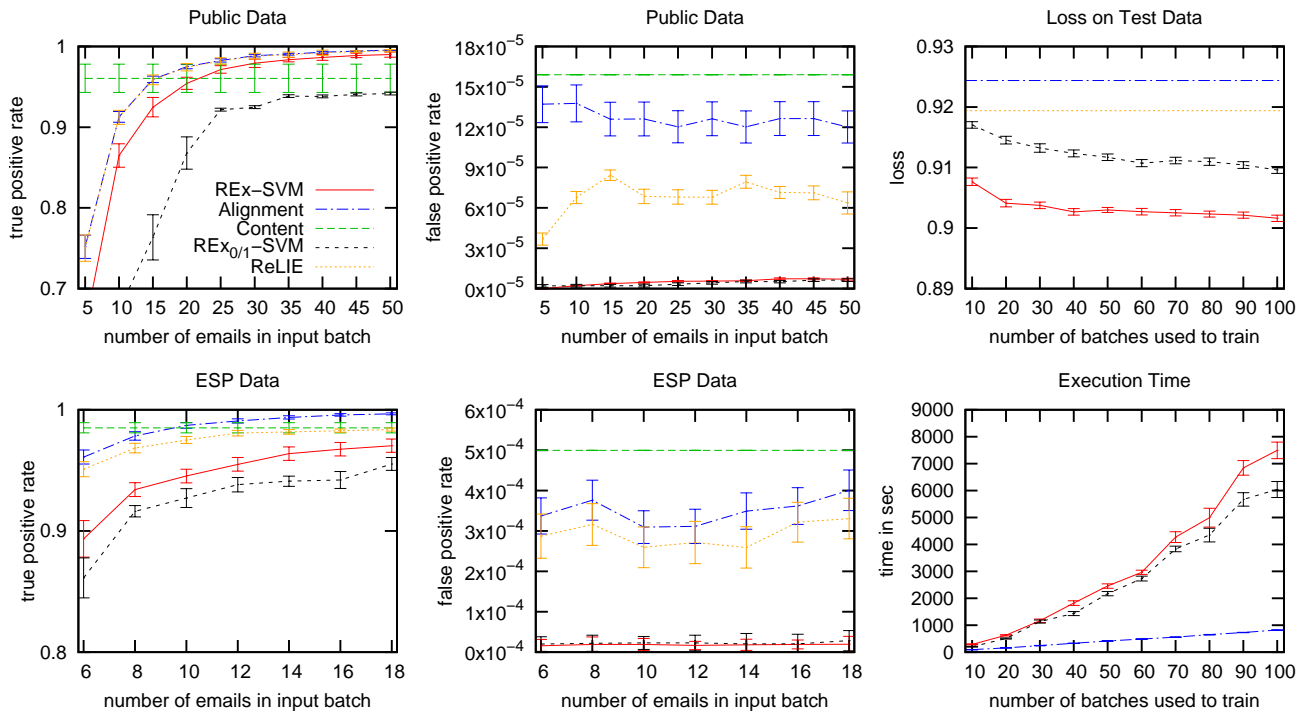

Figure 5. Empirical results on public and ESP data sets. Error bars indicate standard errors.

### 5.2. Spam Filtering Performance

We evaluate the ability of *REx-SVM* and baselines to identify the exact extension of email campaigns. We use the *alignment* of the strings in **x** as a baseline. In addition, *ReLIE* (Li et al., 2008) searches for a regular expression that matches the emails in the input batch and does not match any of the additional negative examples by applying a set of transformation rules; we use the alignment of the input batch as starting point. *ReLIE* receives an additional 10,000 emails that are not part of any batch as negative data. An additional *content*-based filter employed by the provider has been trained on several million spam and non-spam emails.

In order to be able to measure false-positive rates (the rate at which emails that are not part of a campaign are erroneously included), we combine the ESP data set with an additional 135,000 non-spam emails, also from the provider. Additionally, we use a *public* data set that consists of 100 batches of emails extracted from the *Bruce Guenther archive*[1], containing a total of 63,512 emails. To measure false-positive rates, we combine this collection of spam batches with 17,419 emails from the *Enron corpus*[2] of non-spam emails and 76,466 non-spam emails of the *TREC corpus*[3]. The public data set is available to researchers.

In an outer loop of leave-one-out cross validation, one batch is held back to evaluate the true-positive rate (the proportion of the campaign that is correctly recognized). In an inner loop of 10-fold cross validation, regularization parameter $C$ is tuned.

Figure 5 shows the true and false positive rates for all methods and both data sets. The horizontal axis displays the number of emails in the input batch **x**. Error bars indicate the standard error. The *alignment* exhibits the highest true-positive rate and a high false-positive rate because it is the most-general bound of the decoder's search space. *ReLIE* needs only very few or zero replacement steps until no negative examples are covered. Consequently, it has similarly high true- and false-positive rates. *REx-SVM* attains a slightly lower true positive rate, and a substantially lower false-positive rate. The false-positive rates of *REx* and $REx_{0/1}$ lie more than an order of magnitude below the rate of the commercial *content*-based spam filter employed by the email service provider. The zero-one loss leads to comparable false-positive but lower true-positive rates, rendering the loss func-tion of Equation 1 preferable to the zero-one loss.

---

[1] http://untroubled.org/spam/

[2] http://www.cs.cmu.edu/~enron/
[3] http://trec.nist.gov/data/spam.html

The execution time to learn a model (bottom right) is consistent with prior findings of between linear and quadratic for the SVM optimization process.

## 6. Related Work

Gold (1967) shows that it is impossible to exactly identify any regular language from finitely many positive examples. Our notion of minimizing an expected difference between conjecture and target language over a distribution of input strings reflects a more statistically-inspired notion of learning. Also, in our problem setting the learner has access to pairs of sets of strings and corresponding regular expressions.

Most work of identification of regular languages focuses on learning automata (Denis, 2001; Clark & Thollard, 2004). While these problems are identical in theory, transforming generated automata into regular expressions can lead to lengthy terms that do not lend themselves to human comprehension (Fernau, 2009). Some work focuses on restricted classes, such as expressions in which each symbol occurs at most $k$ times (Bex et al., 2008), disjunction-free expressions (Brāzma, 1993), and disjunctions of left-aligned disjunction-free expressions (Fernau, 2009).

Xie et al. (2008) use regular expressions to detect URLs in spam batches and develop a spam filter with low false positive rate. The *ReLIE*-algorithm (Li et al., 2008) (used as a reference method in our experiments) learns regular expressions from positive and negative examples given an initial expression by applying a set of transformation rules as long as this improves the separation of positive and negative examples.

## 7. Conclusions

Complementing the language-identification paradigm, we pose the problem of learning to map a set of strings to a target regular expression. Training data consists of batches of strings and corresponding expressions. We phrase this problem as a learning problem with structured output spaces and engineer an appropriately loss function. We derive the resulting optimization problem, and devise a decoder that searches a space of specializations of a maximal alignment.

From our case study we conclude that *REx-SVM* gives a high true positive rate at a false positive rate that is more than an order of magnitude lower than that of a commercial *content*-based filter. The system is being used by a commercial email service provider and complements *content*-based and IP-address based filtering.

## Acknowledgments

This work was funded by a grant from STRATO AG.